\documentclass[runningheads]{llncs}
\usepackage[T1]{fontenc}

\usepackage{graphicx}
\usepackage{amsmath}
\usepackage{booktabs}
\usepackage{multirow}
\usepackage{lipsum}
\usepackage[graphicx]{realboxes}
\usepackage{rotating}
\usepackage[misc,geometry]{ifsym}
\begin{document}
\title{CLST: \underline{C}old-Start Mitigation in Knowledge Tracing by Aligning a Generative \underline{L}anguage Model as a \underline{S}tudents’ Knowledge \underline{T}racer}
\titlerunning{Cold-start Mitigation in Knowledge Tracing by Aligning a Language Model}
%
\author{Heeseok Jung\inst{1} \and
Jaesang Yoo\inst{1} \and
Yohaan Yoon\inst{1} \and
Yeonju Jang\inst{2}\textsuperscript{(\Letter)}}
\authorrunning{Jung et al.}
%
\institute{Classting Edutech Research, Seoul, Republic of Korea 
\email{\{poco2889,jsyoo,eddie\}@classting.com}\\\and
Korea University, Seoul, Republic of Korea\\
\email{spring0425@korea.ac.kr}}
\maketitle              
\begin{abstract}
Knowledge tracing (KT), wherein students’ problem-solving histories are used to estimate their current levels of knowledge, has attracted significant interest from researchers. However, most existing KT models were developed with an ID-based paradigm, which exhibits limitations in cold-start performance. These limitations can be mitigated by leveraging the vast quantities of external knowledge possessed by generative large language models (LLMs). In this study, we propose cold-start mitigation in knowledge tracing by aligning a generative language model as a students’ knowledge tracer (CLST) as a framework that utilizes a generative LLM as a knowledge tracer. Upon collecting data from math, social studies, and science subjects, we framed the KT task as a natural language processing task, wherein problem-solving data are expressed in natural language, and fine-tuned the generative LLM using the formatted KT dataset. Subsequently, we evaluated the performance of the CLST in situations of data scarcity using various baseline models for comparison. The results indicate that the CLST significantly enhanced performance with a dataset of fewer than 100 students in terms of prediction, reliability, and cross-domain generalization.

\end{abstract}
\section{Introduction}
As we enter the era of digital transformation, online learning is becoming increasingly prevalent worldwide. In this context, intelligent tutoring systems (ITSs) have become increasingly significant in the field of education \cite{abdelrahman2023knowledge}. These systems can significantly enhance student learning by utilizing technologies that predict academic achievement and dropout rates \cite{emerson2023early,wang2023log,colpo2024lessons}.

Among the various technologies used in ITSs, knowledge tracing (KT) has emerged as a key component that enables personalized learning \cite{shen2021learning}. To predict students' responses to new exercises, KT models their knowledge states by analyzing their past performance \cite{wu2023fusing}. Understanding the students’ knowledge states helps in recommending appropriate learning materials and offering relevant feedback, thereby enhancing learning outcomes.

A variety of KT models have been proposed to meet these expectations, including both traditional and deep learning (DL)-based models \cite{abdelrahman2023knowledge}. In particular, the performance of DL-based KT models in terms of predicting knowledge states has notably improved with successive developments.

The majority of existing KT models are dominated by the ID-based paradigm. Although such ID-based models often perform well in predicting students’ knowledge states, they share one limitation: prediction performance is significantly poor in cold-start situations \cite{zhao2020cold}. As an issue stemming from scarce data on users or items in recommendation systems \cite{sahebi2011community}, the cold-start problem is regarded as an important issue in the KT field, occurring in the presence of insufficient data pertaining to certain students or problem-solving interactions \cite{wu2022sgkt}. When an educational institution or EdTech company that operates an online learning platform attempts to develop an ITS using a KT model for the first time, it is likely to encounter cold-start problems in the early stages. Consequently, educational institutions may encounter challenges in utilizing most existing KT models during the development of new ITSs.

Many studies have made efforts to alleviate the cold-start issue inherent to KT, primarily by utilizing various types of side information such as the language proficiency of learners \cite{jung2023language}, difficulty level of questions \cite{liu2020improving,zhang2021multi}, and students' question-answering patterns \cite{xu2023improving}. However, these methods require effort to obtain additional information beyond each learner’s problem-solving history, which may be difficult in the early stages of a service.

On the other hand, generative large language models (LLMs) demonstrate outstanding performance in a wide range of fields and downstream applications \cite{zhou2023survey,jin2023large,li2023large}. Generative LLMs, such as GPT, are auto-regressive language models that can generate text similar to human writing. These models contain large quantities of external knowledge and are able to extract high-quality textual features.

The development of KT models using external knowledge inherently possessed by generative LLMs opens the possibility of improving performance in cold-start situations as well as enhancing the domain adaptability of KT models. In a cold-start scenario, the scarcity of information from the target domain can be supplemented by using information from other domains as training data \cite{cheng2022adaptkt}. However, domain adaptation is challenging because most ID-based KT models currently in use are domain-specific. If a generative LLM is used to train a KT model with diverse domains, high performance can be achieved even in a small number of target domains owned by educational institutions.

Despite the potential of this approach, little research has been conducted on the applicability of generative LLMs to KT tasks. One such study was conducted \cite{neshaei2024towards}; however, this attempt failed to fully utilize the strengths of generative LLMs, only expressing exercise description information through IDs. Furthermore, their model relied solely on the number of correct and incorrect answers submitted by each learner, whereas it is also important to account for the order in which each learner solves problems. As such, research on modeling KT through language processing remains insufficient. Given their capabilities, generative LLMs can be deployed to overcome the limitations of the ID-based paradigm. As a result, more research is needed to determine the efficacy of expressing KT in natural language using generative LLMs.

In this study, we propose a framework called “cold-start mitigation in KT by aligning a generative language model as a students’ knowledge tracer (CLST).” We conducted experiments to determine whether a generative LLM aligned with KT could perform well in cold-start scenarios. Initially, we collected data for a variety of subjects, including mathematics, social studies, and science. The KT task was then framed as a natural language processing (NLP) task by expressing problem-solving data in natural language, and the formatted KT dataset was used to fine-tune the generative LLM. Finally, multiple experiments were conducted to investigate CLST's performance in cold-start situations. This study was conducted to address the following research questions (RQs):

\begin{itemize}
  \item RQ1: In cold-start scenarios, does the proposed method demonstrate successful prediction performance?
  \item RQ2: How can exercises be effectively represented in natural language?
  \item RQ3: Does the proposed model provide a convincing prediction of learners’ knowledge states?
  \item RQ4: Is the proposed model effective at predicting student knowledge in cross-domain scenarios?
\end{itemize}

The remainder of this paper is organized as follows: A literature review on KT and generative LLMs in personalized learning are presented in Section 2. Section 3 describes the methodology developed in this study. Section 4 presents and discusses the experimental results. Finally, Section 5 concludes the study.

\section{Literature Review}
\subsubsection{2.1 Knowledge tracing\\}

KT utilizes students' problem-solving histories to approximate their current knowledge states and subsequently predict their future responses. A student’s problem-solving history can be represented as $X=\{(e_1,y_1),(e_2,y_2),\cdots,(e_t,y_t)\}$, where $e_t$ corresponds to an exercise that has been solved at a specific time $t$, and $y_t$ represents the correctness of the student’s response to $e_t$.

KT models can be distinctly classified as traditional or DL-based \cite{abdelrahman2023knowledge}. Traditional approaches of KT include bayesian knowledge tracing (BKT) and factor analysis models. BKT \cite{corbett1994knowledge} is a Markov-process-based approach that represents each student’s knowledge state as a collection of binary values, whereas factor analysis approaches analyze the factors that affect students’ knowledge states, such as attempt counts and exercise difficulty. Performance factor analysis (PFA) \cite{pavlik2009performance} and knowledge tracing machines (KTM) \cite{vie2019knowledge} are representative works on factor analysis.

The growth of online education has led to a substantial accumulation of problem-solving data, allowing DL-based KT models to reach outstanding performance. Such methods include recurrent neural network (RNN)-based \cite{piech2015deep,yeung2018addressing}, memory-augmented neural network (MANN)-based \cite{zhang2017dynamic,abdelrahman2019knowledge,kim2021dikt}, transformer-based \cite{pandey2019self,ghosh2020context,yin2023tracing}, and graph neural network (GNN)-based \cite{nakagawa2019graph,ni2023hhskt,yang2024heterogeneous} models. In this study, the BKT \cite{corbett1994knowledge} and factor analysis models \cite{pavlik2009performance,van1997handbook} were chosen from the traditional KT models, and the RNN-based model \cite{piech2015deep}, MANN-based model \cite{zhang2017dynamic}, and transformer-based model \cite{ghosh2020context} were chosen as baseline models and used in the experiment.

\subsubsection{2.2 Generative LLMs in personalized learning\\}

Generative LLMs are pre-trained auto-regressive language models, such as GPT-3, capable of generating human-like text \cite{peng2023study}. These probabilistic models serve as the foundation for natural language processing (NLP) techniques, which enable the processing of natural language using algorithms. The term ‘large’ denotes the extensive number of parameters each of these models contains, while the term ‘generative’ refers to a subset of LLMs designed for text generation. Generative LLMs are now at the core of a variety of applications, including summarization and translation, generally delivered through dialogue-like communication with the user \cite{zhao2023survey}.
 
In light of their massive potential, attempts are being made to utilize generative LLMs in the field of education, with ongoing studies actively exploring methods to facilitate personalized learning through the use of these models. To date, LLMs have exhibited effective performance in generating various educational materials, including multiple-choice questions \cite{olney2023generating}, stories for reading comprehension assessments \cite{bulut2022automatic}, and quizzes \cite{dijkstra2022reading}. Additionally, studies have been carried out on adaptive curriculum design \cite{sridhar2023harnessing}, automatic grading \cite{malik2019generative}, and automated feedback generation for assignments \cite{abdelghani2022gpt} utilizing generative LLMs.
 
Although several studies have been conducted on the use of generative LLMs in personalized learning, few attempts have been made to apply them to the KT field, which is a core technology to support personalized learning. The aforementioned study conducted by \cite{neshaei2024towards} was limited in that it did not fully leverage the capabilities of generative LLMs, as exercises were simply represented using IDs. Moreover, the sequence of problem-solving by learners is crucial in the KT task, whereas the study in question only considered the number of correct and incorrect answers by each learner.
 
Studies on KT have been conducted in an NLP context, leveraging textual data from questions. For instance, exercise-enhanced recurrent neural network (EERNN) \cite{su2018exercise} leverages the text of each exercise to predict student responses utilizing the Markov property or attention mechanism. Word2vec \cite{mikolov2013distributed} is used to transform the words of each exercise into word embeddings, and exercise embeddings are subsequently learned by a bidirectional long short-term memory (LSTM) network. Exercise-aware knowledge tracing (EKT) \cite{liu2019ekt} is an extended version of EERNN, where a memory network is deployed to quantify the contribution of each exercise to student mastery on multiple knowledge components (KCs). Relation-aware self-attention model for knowledge tracing (RKT) \cite{pandey2020rkt} leverages interaction data and the textual content of each exercise to compute an exercise relation matrix for KT. Furthermore, \cite{tong2020hgkt} employed the BERT \cite{devlin2018bert} embedding similarities between textual data from exercises to construct a graph, thereby predicting student responses using a hierarchical GNN and two attention mechanisms.

However, these methods did not rely on generative language models, instead solely utilizing NLP to obtain exercise representations. In other words, there is still plenty of potential for further research on the effectiveness of generative-LLM-based KT models. In this study, we examined the efficacy of aligning generative LLMs to KT by expressing the KT task in natural language.

\section{Methodology}
\subsubsection{3.1 Problem formulation\\} Letting $E$ be a set of exercises. A studnet's responses to exercises in $E$, and records of one's problem-solving can be represented as the set $X_{1:t}=\{x_1,x_2,\cdots,x_t\}$, comprising tuples $x_t=(e_t,y_t)$. Here, $e_t\in E$ represents the exercise solved at a specific time step $t$, and $y_t \in \{0,1\}$ represents the correctness of the answer recorded by the student at $t$. With this notation, we can formulate the KT task in the following way: Given a student’s historical information $X_{1:t}$, KT aims to predict the probability of correctly solving a new exercise at $t+1$. In the following equation, $f$ denotes the KT model.
\begin{equation}
\hat{y}_{t+1}=f(X_{1:t}, e_{t+1})
\end{equation}

\subsubsection{3.2 CLST: Cold-start mitigation in KT by aligning a generative language model as a students’ knowledge tracer\\}
\paragraph{3.2.1 KTLP formatting: Utilizing natural language as KT information carrier\\}
A student’s problem-solving history X must be transformed into textual sequences $X^{text}$ using prompt templates for generative LLMs. We start by organizing ‘Task Input’ to instruct the model to determine whether the student will solve the target exercise based on their historical interactions, and then output a binary response of “Yes” or “No”.  Here, each exercise is represented by a textual description (e.g., name of KC), with interactions represented as tuples of (exercise description, correctness), connected by the ‘->’ symbol.

Additionally, the binary label $y\in{0,1}$ is converted into a binary key answer word $y^{text}\in\{yes,\ no\}$ to organize the ‘Task Output’. The aforementioned process allows us to frame the KT task as a language processing task. We call this process KT as language processing (KTLP) formatting, which can be formulated as follows:
\begin{equation}
{X_{1:t}}^{text}=g(X_{1:t},e_{t+1})\ ,\ {y_{t+1}}^{text}=g(y_{t+1})
\end{equation}

Here, $g$ denotes the KTLP formatting function. Figure 1 depicts a sample KTLP-formatted interaction.

\begin{figure}
  \centering
  \rotatebox{0}{
    \begin{minipage}{1\textwidth}
      \includegraphics[width=1\textwidth]{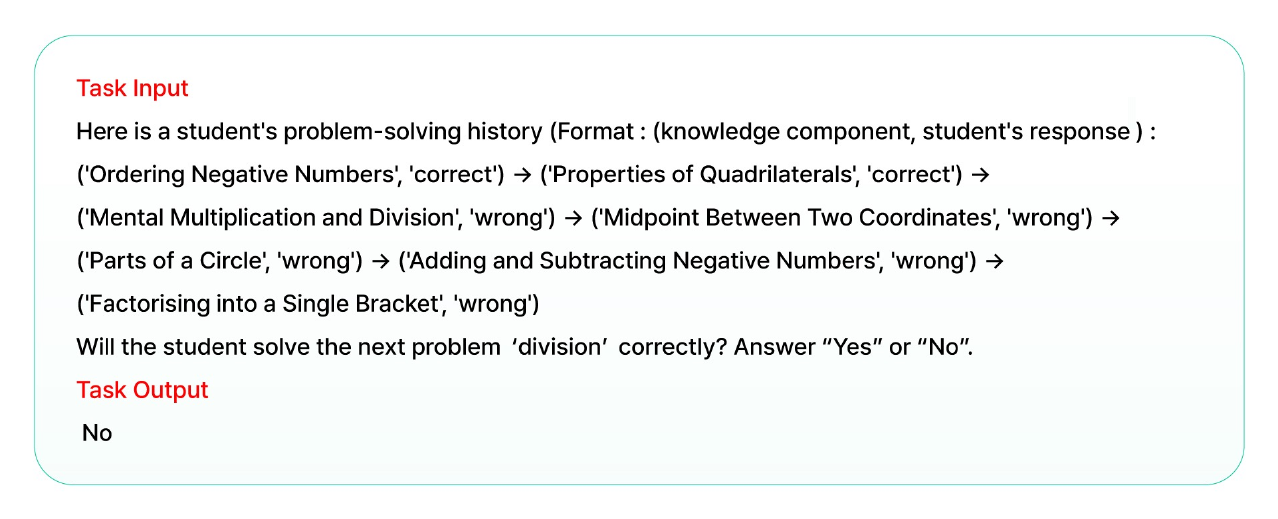}
      \caption{Sample KTLP-formatted interaction.}
    \end{minipage}}
  \label{figure1}
\end{figure}

\paragraph{3.2.2 Response prediction with CLST\\}
Upon taking the discrete tokens of ${X_{1:t}}^{text}$ as input, an LLM is used to generate the next token ${{\hat{y}}_{t+1}}^{text}$ as the output. This process can be formulated as follows:
\begin{equation}
l_{t+1}=M({X_{1:t}}^{text})\in R^V
\end{equation}
\begin{equation}
p_{t+1}=\ Softmax(l_{t+1})\in R^V
\end{equation}
\begin{equation}
{{\hat{y}}_{t+1}}^{\ text}\sim p_{t+1}
\end{equation}

where $M$ denotes the LLM, $V$ denotes the vocabulary size, and ${{\hat{y}}_{t+1}}^{text}$ denotes the next predicted token drawn from the probability distribution $p_{t+1}$.

However, the KT model’s output should be a floating-point number ${\hat{y}}_{t+1}\in\left[0,1\right]$ rather than a discrete token ${{\hat{y}}_{t+1}}^{text}$. Therefore, following prior studies on point-wise scoring tasks (e.g., recommendation), we take the logits $l_{t+1}\in R^V$ and apply a bi-dimensional softmax over the logits of binary answer words. The KT prediction of LLM can thus be written as:
\begin{equation}
{\hat{y}}_{t+1}=\frac{exp\left(l_{t+1,\ yes}\right)}{exp\left(l_{t+1,yes}\right)+exp\left(l_{t+1,no}\right)}
\end{equation}

\paragraph{3.2.3  Instruction fine-tuning using KTLP-formatted data with LoRA architecture\\}
Since we generated both the task input and the task output in natural language, we can optimize the LLM by following the common instruction fine-tuning and causal language modeling paradigm. This process can be formulated as follows:
\begin{equation}
    \max_{\Phi}\sum_{(x,y)\in Z}\sum_{k=1}^{|y|}{{log}(P_\Phi (y_k|x,y_{<k}))}
\end{equation}

Here, $x$ and $y$ denote the KTLP-formatted input and output, respectively, $Z$ denotes the training dataset, $y_k$ represents the $k$-th token of $y$, $y_{<k}$ denotes the token preceding $y_k$, and $\Phi$ represents the parameters of $M$.

However, it is highly resource-intensive to fine-tune every parameter of an LLM. To utilize training resources efficiently, we instead adopted low-rank adaptation (LoRA) \cite{hu2021lora}. As a parameter-efficient fine-tuning method, LoRA incorporates smaller trainable matrices for each layer of the model through low-rank decomposition and the freezing of pre-trained model parameters, thereby enabling lightweight tuning. As a result, the original parameters can be preserved in a frozen state while additional information in the fine-tuning dataset is efficiently incorporated by training smaller matrices. The final fine-tuning process can be formulated as follows:

\begin{equation}
	\max_{\Theta}\sum_{(x,y)\in Z}\sum_{k=1}^{|y|}{{log}(P_{\Phi+\Theta}(y_k|x,y_{<k}))}
\end{equation}
Here, $\Theta$ represents the LoRA matrices, which are only updated during the fine-tuning process.

\subsubsection{3.3 Dataset\\}

To investigate the effectiveness of CLST, we collected data pertaining to mathematics, social studies, and science subjects. Table 1 lists detailed statistics for each dataset.

For mathematics data, we selected the open benchmark datasets NIPS34, Algebra05, and Assistments09 (Assist09), which are used as standard benchmarks for KT methods \cite{sun2024progressive,yin2023tracing}. The NIPS34 dataset was sourced from the third and fourth tasks of the NeurIPS 2020 Education Challenge. Gathered from the Eedi platform, the data encompass students’ problem-solving histories for multiple-choice math exercises \cite{wang2020instructions}. The Algebra05 dataset was sourced from the EDM challenge of the 2010 KDD Cup, comprising the responses of 13- to 14-year-old students to algebra exercises \cite{stamper2010}. The Assistments09 dataset contains student response data on math exercises collected from the ASSISTments platform in 2009–2010 \cite{feng2009addressing}.

For social studies and science, we gathered data from Classting AI Learning, an online education platform that offers learning materials in a range of academic subjects for K–12 students.

\begin{table}[]
\caption{Detailed statistics of datasets used in this study.}
\label{tab1}
\begin{tabular}{llllll}
\cline{1-6}
                                  & NIPS34    & Algebra05 & Assist09 & Social   studies & Science   \\ \cline{1-6}
\#   interaction                  & 1,382,727 & 607,014   & 282,071  & 4,215,373        & 5,484,647 \\
\#   learners                     & 4,918     & 571       & 3,644    & 40,217           & 45,211    \\
\#   exercises                    & 948       & 173,113   & 17,727   & 14,168           & 29,795    \\ 
\# KCs                            & 57        & 112       & 123      & 40               & 76        \\
Median   interactions per learner & 239       & 580       & 25       & 40               & 35        \\
Median   KCs per learner          & 26        & 54        & 5        & 3                & 3        \\
\cline{1-6}
\end{tabular}
\end{table}

\subsubsection{3.4 Baseline models\\}
In our comparative experiments, traditional, DL-based, and NLP-enhanced DL-based KT models were selected as baselines.

As traditional KT models, we selected the BKT, item response theory (IRT), and PFA models. BKT is a hidden Markov model that encodes learners' knowledge states with binary variables \cite{corbett1994knowledge}. IRT models KT via logistic regression, accounting for each student’s ability as well as the difficulty of each exercise \cite{van1997handbook}. PFA is another logistic regression model that incorporates exercise difficulty alongside each student’s prior successes and failures \cite{pavlik2009performance}. 

As DL-based KT models, we selected deep knowledge tracing (DKT), dynamic key-value memory networks for knowledge tracing (DKVMN), and attentive knowledge tracing (AKT). DKT \cite{piech2015deep} is a standard RNN-based KT model that predicts students’ knowledge states with a single LSTM layer \cite{shen2024survey}. DKVMN \cite{zhang2017dynamic} is a KT model that uses static memory to encode latent KCs and dynamic memory to track student proficiency for the latent KCs \cite{shen2024survey}.  AKT \cite{ghosh2020context} is an attention-mechanism-based KT model that improves predictive performance by incorporating context-aware attention with Rasch model-based encoding.

We implemented NLP-enhanced DL-based KT models. These models initialize exercise embedding vectors in DL-based KT models with encoded text features of the KC name that corresponds to each exercise. which covers each exercise Specifically, we used OpenAI\footnotemark[1]\footnotetext[1]{https://platform.openai.com/docs/guides/embeddings}’s \textit{text-embedding-3-large} to encode text features, with the resulting KT models denoted as $\text{DKT}_{text}$, $\text{DKVMN}_{text}$, and $\text{AKT}_{text}$.

\subsubsection{3.5 Experimental settings\\}

To guarantee the quality of the data, only interactions from students who responded to more than five exercises were used.

Because the objective of this study was to mitigate problems that arise from cold-start situations, we alternately limited the number of learners in the training set to 64, 32, 16, and 8. We executed each method five times with different random seeds and reported the average outcomes. For the mathematics data, 20\% of the learners were held out for the test set. For the social studies and science data, we selected 1000 random learners as the test set for each subject.

The selection of a base model requires careful consideration. Among existing generative LLMs, many do not provide accessibility to their model weights or APIs. Furthermore, data security is a critical concern in the field of education, necessitating additional discussion about the use of third-party APIs (e.g., OpenAI). After careful consideration, we selected the instruction-tuned Mistral-7B\footnotemark[2]\footnotetext[2]{https://huggingface.co/mistralai/Mistral-7B-Instruct-v0.2} \cite{jiang2023mistral}.

Because it was essential to evaluate predictive performance with limited data, we only considered the first 50 interactions per student in the experiment \cite{wang2021temporal}. For other baseline hyperparameters, we adhered to the settings described by \cite{wang2021temporal,lee2022contrastive}, as their evaluation protocols were similar to our own.

\section{Results and discussion}
To evaluate the effectiveness of CLST from multiple perspectives, as well as address the four RQs, we conducted four experiments with the following objectives: 1) compare predictive performance between CLST and the baseline models in cold-start scenarios; 2) investigate the effectiveness of each component of the CLST through an ablation study; 3) analyze learning trajectories using CLST; and 4) compare predictive performance between CLST and the baseline models in cross-domain tasks. The following subsections discuss our experimental results.

\subsubsection{4.1 Predictive performance under cold-start scenarios\\}

\begin{figure}
  \centering
  \rotatebox{0}{
    \begin{minipage}{1\textwidth}
    \centering
      \includegraphics[width=1\textwidth]{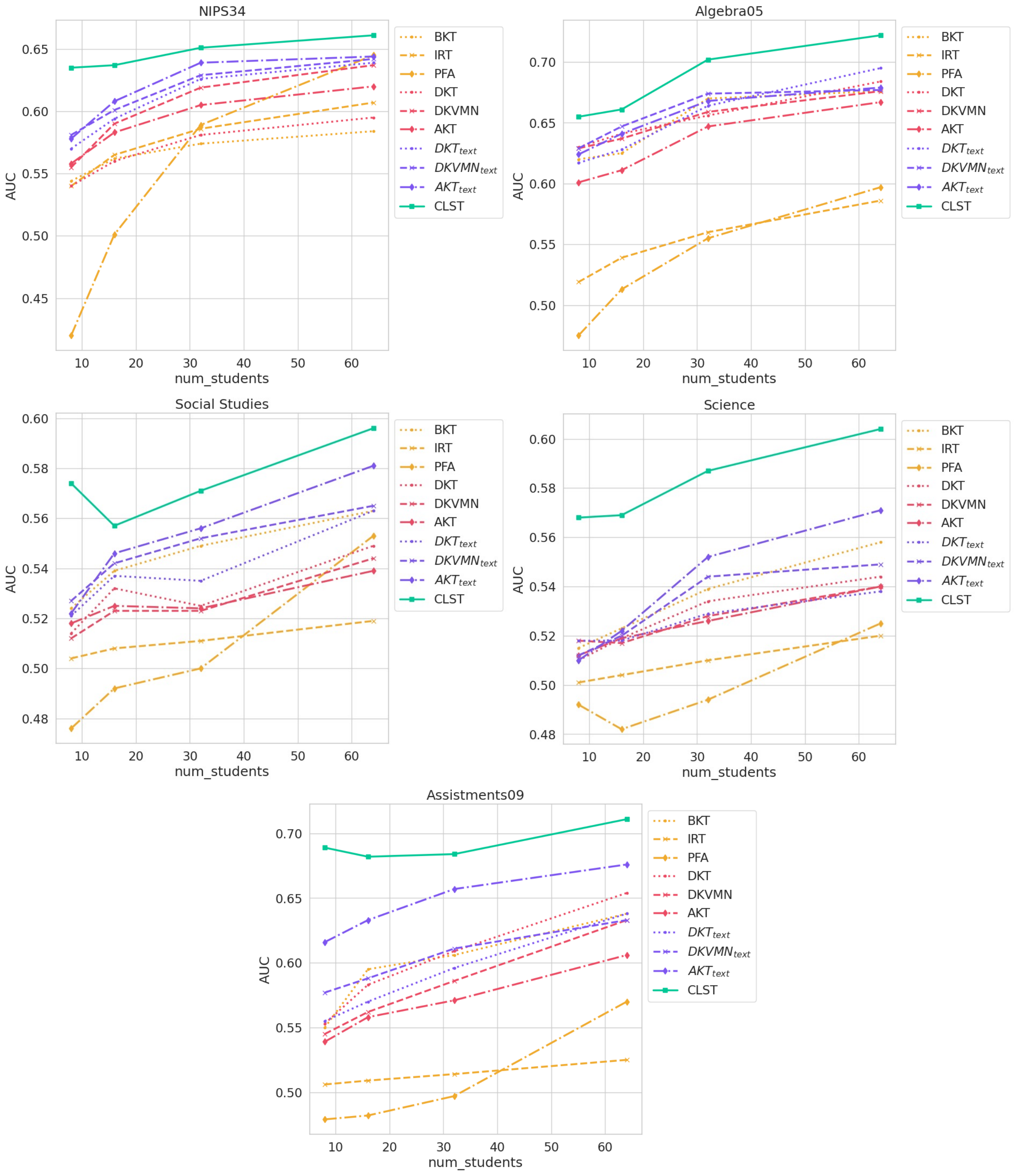}
      \caption{Cold-start performance on each dataset.}
    \end{minipage}}
  \label{fig2}
\end{figure}

We evaluated the predictive performance of CLST in cold-start scenarios involving a limited number of students, with the training set sequentially reduced from 64 to 8 students. All baseline models listed in Section 3.4 were chosen for comparison in this experiment. We measured the area under the receiving operator characteristic (AUC) as an evaluation metric. AUC is the most widely used indicator for evaluating KT performance, with higher scores indicating better performance \cite{liu2022pykt}.

The experimental results are presented in Figure 2, where traditional, DL-based, and NLP-enhanced models are represented by yellow, red, and purple curves, respectively, whereas CLST is represented by a green curve.

As observed from the figure, CLST outperformed every baseline model, irrespective of the size of the training set. Furthermore, models utilizing NLP demonstrated superior overall performance in comparison to traditional and DL-based methods.

Table 2 presents detailed quantitative results of the cold-start experiments, where the best AUC score for each trial is bolded and underlined, while the second-highest AUC score is underlined. CLST outperformed the second-best model by up to 24.52\%, 14.66\%, 12.31\%, and 8.71\% for a training set containing 8, 16, 32, and 64 students, respectively. Furthermore, when comparing performance between CLST and the baseline models across different datasets, CLST exhibited improvements of up to 13.69\% on the NIPS34 dataset, 5.62\% on the Algebra05 dataset, 24.52\% on the Assist09 dataset, 9.62\% on the Classting Social Studies dataset, and 11.82\% on the Classting Science dataset.

\begin{table}[]
\rotatebox{90}{
\begin{minipage}{1.5 \textwidth}
\label{tab2}
\caption{Detailed results of cold-start experiments.}
\begin{tabular}{|c|c|cccccccccc|}
\hline
\multirow{3}{*}{Dataset} & \multirow{3}{*}{\begin{tabular}[c]{@{}c@{}}\# \\ Students\end{tabular}} & \multicolumn{10}{c|}{Average AUC} \\ \cline{3-12} 
 &  & \multicolumn{3}{c|}{Traditional} & \multicolumn{3}{c|}{DL-based} & \multicolumn{3}{c|}{NLP-enhanced} &  \\ \cline{3-11}
 &  & \multicolumn{1}{c|}{BKT} & \multicolumn{1}{c|}{IRT} & \multicolumn{1}{c|}{PFA} & \multicolumn{1}{c|}{DKT} & \multicolumn{1}{c|}{DKVMN} & \multicolumn{1}{c|}{AKT} & \multicolumn{1}{c|}{$\text{DKT}_{text}$} & \multicolumn{1}{c|}{$\text{DKVMN}_{text}$} & \multicolumn{1}{c|}{$\text{AKT}_{text}$} & CLST \\ \hline
\multirow{4}{*}{NIPS34} & 8 & \multicolumn{1}{c|}{0.544} & \multicolumn{1}{c|}{0.540} & \multicolumn{1}{c|}{0.420} & \multicolumn{1}{c|}{0.540} & \multicolumn{1}{c|}{0.555} & \multicolumn{1}{c|}{0.558} & \multicolumn{1}{c|}{0.570} & \multicolumn{1}{c|}{\underline{0.581}} & \multicolumn{1}{c|}{0.578} & \begin{tabular}[c]{@{}c@{}}\ \ \ \ \ \underline{\textbf{0.635}}\ \ \ \ \ \ \end{tabular} \\ \cline{2-12} 
 & 16 & \multicolumn{1}{c|}{0.562} & \multicolumn{1}{c|}{0.565} & \multicolumn{1}{c|}{0.501} & \multicolumn{1}{c|}{0.560} & \multicolumn{1}{c|}{0.590} & \multicolumn{1}{c|}{0.583} & \multicolumn{1}{c|}{0.594} & \multicolumn{1}{c|}{0.601} & \multicolumn{1}{c|}{\underline{0.608}} & \underline{\textbf{0.637}} \\ \cline{2-12} 
 & 32 & \multicolumn{1}{c|}{0.574} & \multicolumn{1}{c|}{0.586} & \multicolumn{1}{c|}{0.589} & \multicolumn{1}{c|}{0.581} & \multicolumn{1}{c|}{0.619} & \multicolumn{1}{c|}{0.605} & \multicolumn{1}{c|}{0.626} & \multicolumn{1}{c|}{0.629} & \multicolumn{1}{c|}{\begin{tabular}[c]{@{}c@{}}\ \ \ \ \ \underline{0.639}\ \ \  \ \ \ \end{tabular}} & \underline{\textbf{0.651}} \\ \cline{2-12} 
 & 64 & \multicolumn{1}{c|}{0.584} & \multicolumn{1}{c|}{0.607} & \multicolumn{1}{c|}{\underline{0.645}} & \multicolumn{1}{c|}{0.595} & \multicolumn{1}{c|}{0.637} & \multicolumn{1}{c|}{0.620} & \multicolumn{1}{c|}{0.639} & \multicolumn{1}{c|}{0.642} & \multicolumn{1}{c|}{0.644} & \underline{\textbf{0.661}} \\ \hline
\multirow{4}{*}{Algebra05} & 8 & \multicolumn{1}{c|}{0.620} & \multicolumn{1}{c|}{0.519} & \multicolumn{1}{c|}{0.475} & \multicolumn{1}{c|}{\underline{0.630}} & \multicolumn{1}{c|}{0.629} & \multicolumn{1}{c|}{0.601} & \multicolumn{1}{c|}{0.617} & \multicolumn{1}{c|}{0.629} & \multicolumn{1}{c|}{0.624} & \underline{\textbf{0.655}} \\ \cline{2-12} 
 & 16 & \multicolumn{1}{c|}{0.625} & \multicolumn{1}{c|}{0.539} & \multicolumn{1}{c|}{0.513} & \multicolumn{1}{c|}{0.641} & \multicolumn{1}{c|}{0.637} & \multicolumn{1}{c|}{0.611} & \multicolumn{1}{c|}{0.628} & \multicolumn{1}{c|}{\underline{0.647}} & \multicolumn{1}{c|}{0.641} & \underline{\textbf{0.661}} \\ \cline{2-12} 
 & 32 & \multicolumn{1}{c|}{0.670} & \multicolumn{1}{c|}{0.560} & \multicolumn{1}{c|}{0.555} & \multicolumn{1}{c|}{0.656} & \multicolumn{1}{c|}{0.659} & \multicolumn{1}{c|}{0.647} & \multicolumn{1}{c|}{0.664} & \multicolumn{1}{c|}{\underline{0.674}} & \multicolumn{1}{c|}{0.668} & \underline{\textbf{0.702}} \\ \cline{2-12} 
 & 64 & \multicolumn{1}{c|}{0.676} & \multicolumn{1}{c|}{0.586} & \multicolumn{1}{c|}{0.597} & \multicolumn{1}{c|}{0.684} & \multicolumn{1}{c|}{0.676} & \multicolumn{1}{c|}{0.667} & \multicolumn{1}{c|}{\underline{0.695}} & \multicolumn{1}{c|}{0.677} & \multicolumn{1}{c|}{0.679} & \underline{\textbf{0.722}} \\ \hline
\multirow{4}{*}{Assist09} & 8 & \multicolumn{1}{c|}{0.550} & \multicolumn{1}{c|}{0.506} & \multicolumn{1}{c|}{0.479} & \multicolumn{1}{c|}{0.553} & \multicolumn{1}{c|}{0.545} & \multicolumn{1}{c|}{0.539} & \multicolumn{1}{c|}{0.555} & \multicolumn{1}{c|}{0.577} & \multicolumn{1}{c|}{\underline{0.616}} & \underline{\textbf{0.689}} \\ \cline{2-12} 
 & 16 & \multicolumn{1}{c|}{0.595} & \multicolumn{1}{c|}{0.509} & \multicolumn{1}{c|}{0.482} & \multicolumn{1}{c|}{0.583} & \multicolumn{1}{c|}{0.562} & \multicolumn{1}{c|}{0.558} & \multicolumn{1}{c|}{0.570} & \multicolumn{1}{c|}{0.588} & \multicolumn{1}{c|}{\underline{0.633}} & \underline{\textbf{0.682}} \\ \cline{2-12} 
 & 32 & \multicolumn{1}{c|}{0.606} & \multicolumn{1}{c|}{0.514} & \multicolumn{1}{c|}{0.497} & \multicolumn{1}{c|}{0.609} & \multicolumn{1}{c|}{0.586} & \multicolumn{1}{c|}{0.571} & \multicolumn{1}{c|}{0.596} & \multicolumn{1}{c|}{0.611} & \multicolumn{1}{c|}{\underline{0.657}} & \underline{\textbf{0.684}} \\ \cline{2-12} 
 & 64 & \multicolumn{1}{c|}{0.638} & \multicolumn{1}{c|}{0.525} & \multicolumn{1}{c|}{0.570} & \multicolumn{1}{c|}{0.654} & \multicolumn{1}{c|}{0.633} & \multicolumn{1}{c|}{0.606} & \multicolumn{1}{c|}{0.638} & \multicolumn{1}{c|}{0.633} & \multicolumn{1}{c|}{\underline{0.676}} & \underline{\textbf{0.711}} \\ \hline
\multirow{4}{*}{\begin{tabular}[c]{@{}c@{}}Social \\ Studies\end{tabular}} & 8 & \multicolumn{1}{c|}{0.524} & \multicolumn{1}{c|}{0.504} & \multicolumn{1}{c|}{0.476} & \multicolumn{1}{c|}{0.514} & \multicolumn{1}{c|}{0.512} & \multicolumn{1}{c|}{0.518} & \multicolumn{1}{c|}{0.521} & \multicolumn{1}{c|}{\underline{0.527}} & \multicolumn{1}{c|}{0.522} & \underline{\textbf{0.574}} \\ \cline{2-12} 
 & 16 & \multicolumn{1}{c|}{0.539} & \multicolumn{1}{c|}{0.508} & \multicolumn{1}{c|}{0.492} & \multicolumn{1}{c|}{0.532} & \multicolumn{1}{c|}{0.523} & \multicolumn{1}{c|}{0.525} & \multicolumn{1}{c|}{0.537} & \multicolumn{1}{c|}{0.542} & \multicolumn{1}{c|}{\underline{0.546}} & \underline{\textbf{0.557}} \\ \cline{2-12} 
 & 32 & \multicolumn{1}{c|}{0.549} & \multicolumn{1}{c|}{0.511} & \multicolumn{1}{c|}{0.500} & \multicolumn{1}{c|}{0.525} & \multicolumn{1}{c|}{0.523} & \multicolumn{1}{c|}{0.524} & \multicolumn{1}{c|}{0.535} & \multicolumn{1}{c|}{0.552} & \multicolumn{1}{c|}{\underline{0.556}} & \underline{\textbf{0.571}} \\ \cline{2-12} 
 & 64 & \multicolumn{1}{c|}{0.563} & \multicolumn{1}{c|}{0.519} & \multicolumn{1}{c|}{0.553} & \multicolumn{1}{c|}{0.549} & \multicolumn{1}{c|}{0.544} & \multicolumn{1}{c|}{0.539} & \multicolumn{1}{c|}{0.563} & \multicolumn{1}{c|}{0.565} & \multicolumn{1}{c|}{\underline{0.581}} & \underline{\textbf{0.596}} \\ \hline
\multirow{4}{*}{Science} & 8 & \multicolumn{1}{c|}{0.515} & \multicolumn{1}{c|}{0.501} & \multicolumn{1}{c|}{0.492} & \multicolumn{1}{c|}{0.510} & \multicolumn{1}{c|}{0.518} & \multicolumn{1}{c|}{0.512} & \multicolumn{1}{c|}{\underline{0.518}} & \multicolumn{1}{c|}{0.512} & \multicolumn{1}{c|}{0.510} & \underline{\textbf{0.579}} \\ \cline{2-12} 
 & 16 & \multicolumn{1}{c|}{\underline{0.523}} & \multicolumn{1}{c|}{0.504} & \multicolumn{1}{c|}{0.482} & \multicolumn{1}{c|}{0.519} & \multicolumn{1}{c|}{0.517} & \multicolumn{1}{c|}{0.519} & \multicolumn{1}{c|}{0.518} & \multicolumn{1}{c|}{0.520} & \multicolumn{1}{c|}{0.522} & \underline{\textbf{0.580}} \\ \cline{2-12} 
 & 32 & \multicolumn{1}{c|}{0.539} & \multicolumn{1}{c|}{0.510} & \multicolumn{1}{c|}{0.494} & \multicolumn{1}{c|}{0.534} & \multicolumn{1}{c|}{0.528} & \multicolumn{1}{c|}{0.526} & \multicolumn{1}{c|}{0.529} & \multicolumn{1}{c|}{0.544} & \multicolumn{1}{c|}{\underline{0.552}} & \underline{\textbf{0.587}} \\ \cline{2-12} 
 & 64 & \multicolumn{1}{c|}{0.558} & \multicolumn{1}{c|}{0.520} & \multicolumn{1}{c|}{0.525} & \multicolumn{1}{c|}{0.544} & \multicolumn{1}{c|}{0.540} & \multicolumn{1}{c|}{0.540} & \multicolumn{1}{c|}{0.538} & \multicolumn{1}{c|}{0.549} & \multicolumn{1}{c|}{\underline{0.571}} & \underline{\textbf{0.604}} \\ \hline
\end{tabular}
\end{minipage}}
\end{table}

We can therefore provide the following response to RQ: In all datasets and all cold-start scenarios, the proposed method outperformed the baseline KT models; therefore, we can conclude that the cold-start problem in KT has been mitigated.

The cold-start problem has been a major concern in the field of KT \cite{zhao2020cold,wu2022sgkt}. Previous studies have suggested methods that utilize additional information from the exercise side (e.g., relations among KCs) or the learner side (e.g., language proficiency) to address this issue \cite{jung2023language}. However, these methods require effort to acquire additional information that extends beyond each student’s problem-solving history, which may be difficult to collect during the initial stages of service. In contrast, we suggest a novel approach to mitigating the cold-start issue by conceptualizing KT as language processing and leveraging the capabilities of generative LLMs. 

\subsubsection{4.2 Ablation study\\}

To investigate the effectiveness of each component of CLST in addressing the cold-start issue, we conducted two experiments with the following objectives: (1) comparison of performance with respect to exercise representations (2) to evaluate the effectiveness of fine-tuning on model’s reliability.

\paragraph{4.2.1 Comparison of performance with respect to exercise representations\\}
Most existing KT models represent exercises with their distinctive identity values, or IDs. In the experiments conducted in Section 4.1, all models representing exercises using textual descriptions outperformed those using ID-based representations. In particular, the proposed CLST, which expresses all information in natural language, exhibited the highest performance in each cold-start scenario.

\begin{figure}
\centering
\includegraphics[width=0.75\textwidth]{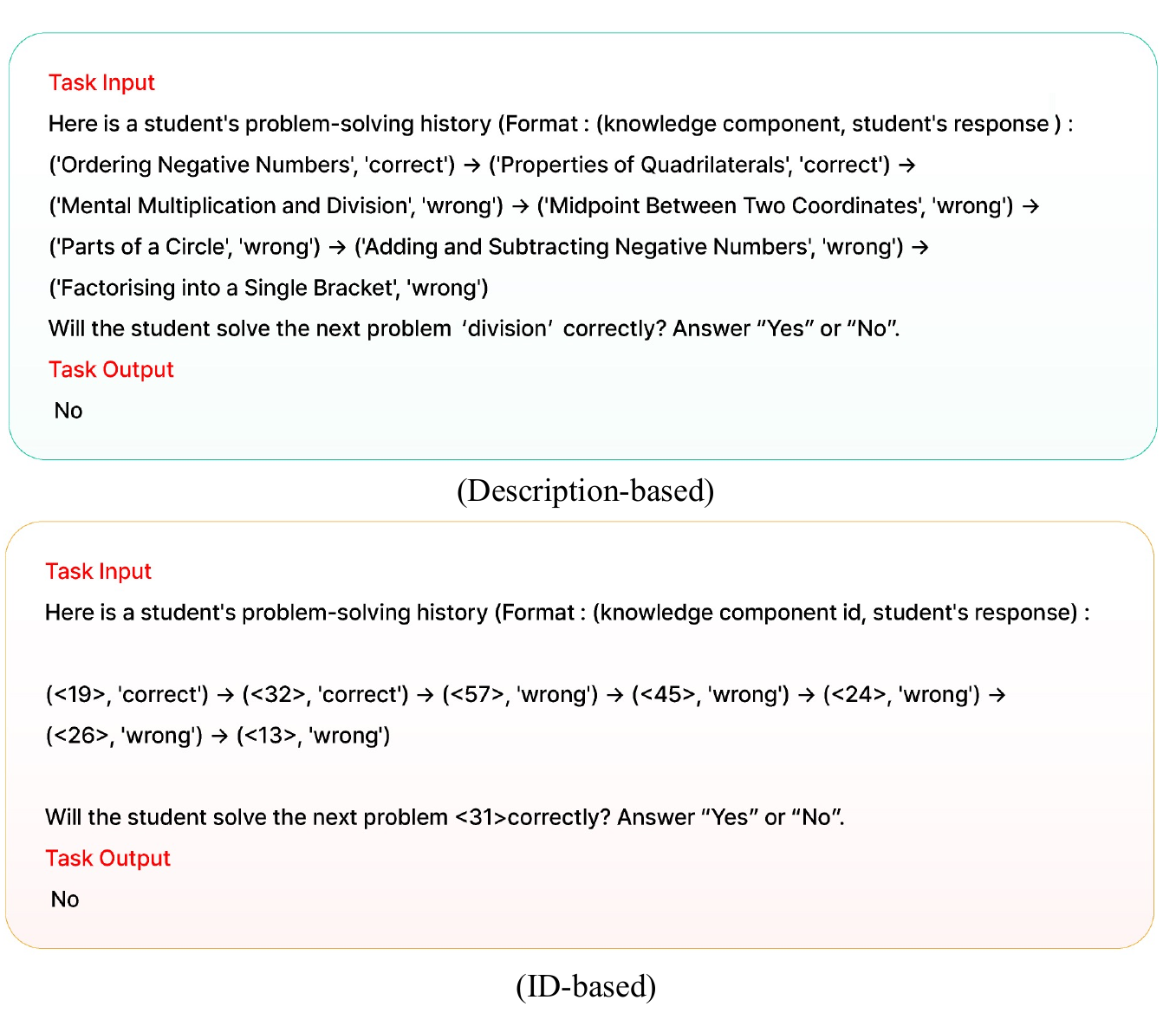}
\caption{Examples of the two methods for representing exercises.} \label{fig3}
\end{figure}

We therefore conducted additional experiments to compare two ways of representing exercises in KTLP format: a description-based method that represents each exercise with the corresponding KC name, which is used in CLST, and the conventional ID-based approach, wherein the ID of the corresponding KC is used to represent each exercise. In this method, the ID of each KC is defined using brackets in the format '<id>' and added as extra vocabulary. Figure 3 presents sample representations of an exercise using both approaches.


\begin{table}[]
\begin{center}
\caption{Comparison of predictive performance (AUC) based on exercise representation method.}
\label{tab3}
\begin{tabular}{|c|c|cccc|}
\hline
\multirow{2}{*}{Dataset} & \multirow{2}{*}{Method} & \multicolumn{4}{c|}{\# Students} \\ \cline{3-6} 
 &  & \multicolumn{1}{c|}{64} & \multicolumn{1}{c|}{32} & \multicolumn{1}{c|}{16} & 8 \\ \hline
\multirow{2}{*}{NIPS34} & ID-based & \multicolumn{1}{c|}{0.658} & \multicolumn{1}{c|}{0.646} & \multicolumn{1}{c|}{\underline{\textbf{0.638}}} & 0.629 \\ \cline{2-6} 
 & Description-based & \multicolumn{1}{c|}{\underline{\textbf{0.661}}} & \multicolumn{1}{c|}{\underline{\textbf{0.651}}} & \multicolumn{1}{c|}{0.637} & \underline{\textbf{0.635}} \\ \hline
\multirow{2}{*}{Algebra05} & ID-based & \multicolumn{1}{c|}{0.659} & \multicolumn{1}{c|}{0.631} & \multicolumn{1}{c|}{0.548} & 0.584 \\ \cline{2-6} 
 & Description-based & \multicolumn{1}{c|}{\underline{\textbf{0.722}}} & \multicolumn{1}{c|}{\underline{\textbf{0.702}}} & \multicolumn{1}{c|}{\underline{\textbf{0.661}}} & \underline{\textbf{0.655}} \\ \hline
\multirow{2}{*}{Assist09} & ID-based & \multicolumn{1}{c|}{0.691} & \multicolumn{1}{c|}{0.661} & \multicolumn{1}{c|}{0.661} & 0.668 \\ \cline{2-6} 
 & Description-based & \multicolumn{1}{c|}{\underline{\textbf{0.711}}} & \multicolumn{1}{c|}{\underline{\textbf{0.684}}} & \multicolumn{1}{c|}{\underline{\textbf{0.682}}} & \underline{\textbf{0.689}} \\ \hline
\multirow{2}{*}{\begin{tabular}[c]{@{}c@{}}Social\\ Studies\end{tabular}} & ID-based & \multicolumn{1}{c|}{\underline{\textbf{0.600}}} & \multicolumn{1}{c|}{0.541} & \multicolumn{1}{c|}{0.537} & 0.567 \\ \cline{2-6} 
 & Description-based & \multicolumn{1}{c|}{0.596} & \multicolumn{1}{c|}{\underline{\textbf{0.571}}} & \multicolumn{1}{c|}{\underline{\textbf{0.557}}} & \underline{\textbf{0.574}} \\ \hline
\multirow{2}{*}{Science} & ID-based & \multicolumn{1}{c|}{0.599} & \multicolumn{1}{c|}{0.584} & \multicolumn{1}{c|}{0.577} & 0.570 \\ \cline{2-6} 
 & Description-based & \multicolumn{1}{c|}{\underline{\textbf{0.604}}} & \multicolumn{1}{c|}{\underline{\textbf{0.587}}} & \multicolumn{1}{c|}{\underline{\textbf{0.580}}} & \underline{\textbf{0.579}} \\ \hline
\end{tabular}
\end{center}
\end{table}

Table 3 presents a comparison of predictive performance with respect to exercise representation. In most cases, the description-based method was observed to outperform the ID-based method.

Based on extensive external knowledge acquired during the pre-training process, the generative LLM is enabled to perform the KT task by considering the relationships between KCs. In general, KT models that account for relationships between KCs are more effective at predicting student performance \cite{chen2018prerequisite,lu2022cmkt}. As a result, when an exercise is represented based on its description, the external knowledge of generative LLMs can be fully utilized, resulting in improved prediction performance. Thus, the results of this experiment can be used to answer RQ2 as follows: When aligning generative LLM with KT, it is more effective to represent each exercise using a description-based method in cold-start situations.

\paragraph{4.2.2 Effectiveness of fine-tuning on model’s reliability\\}
\begin{figure}
  \centering
  \rotatebox{0}{
    \begin{minipage}{1\textwidth}
      \includegraphics[width=1\textwidth]{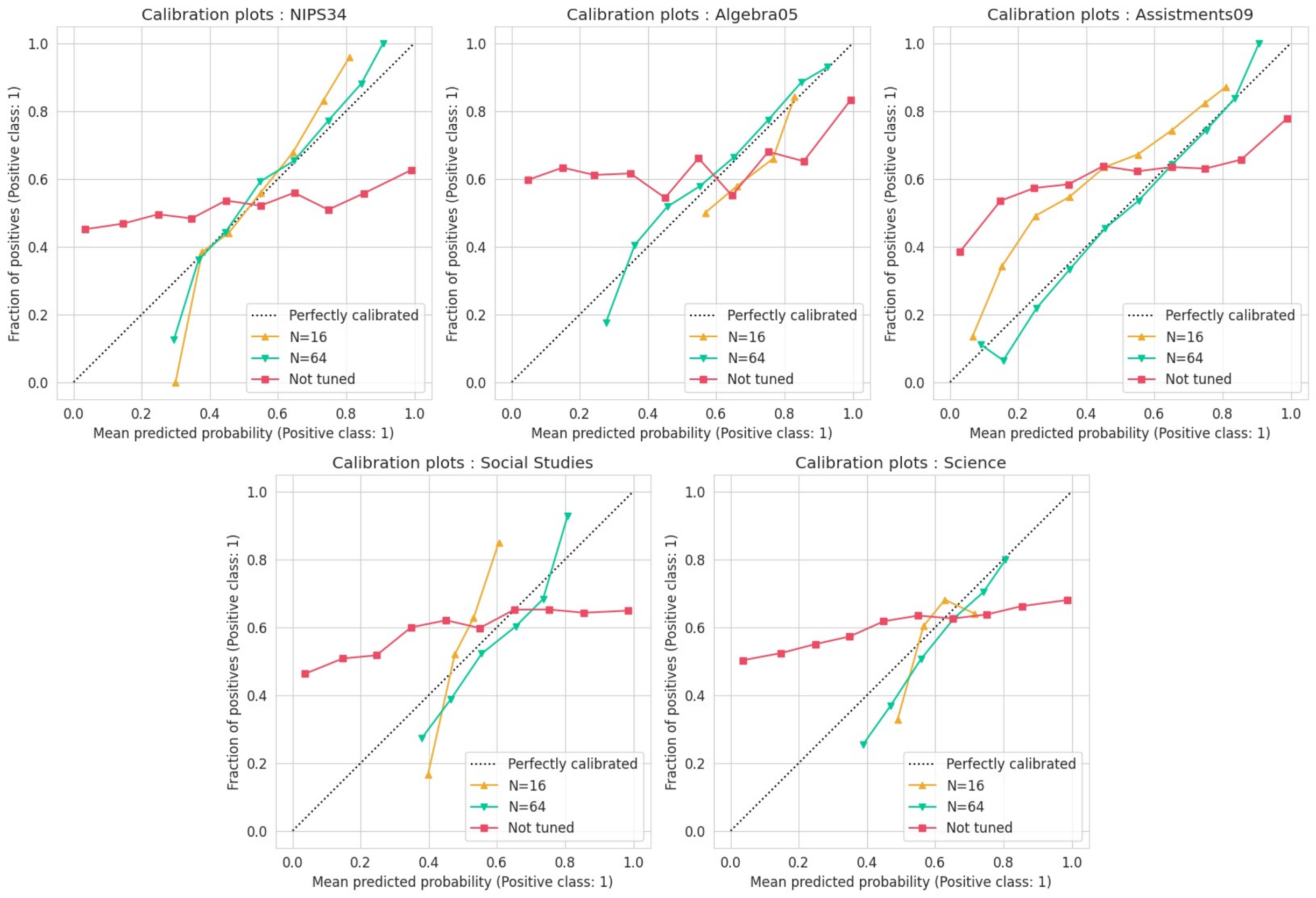}
      \caption{Changes in output calibration as an effect of fine-tuning.}
    \end{minipage}}
  \label{fig4}
\end{figure}

To investigate the effects of fine-tuning with KTLP-formatted data, we compared the calibration plot outputs from an untuned model (\textit{mistral-7b-instruct-v02}, the backbone model selected for this study) and its fine-tuned counterpart.

Figure 4 shows the changes in output calibration as an effect of fine-tuning for each dataset. Each calibration plot is an illustration that compares the average predicted value of a model's output for each bin (x-axis) with the fraction of positive classes (y-axis). A diagonal dotted line in the figure represents the ideal calibration that a model can achieve by predicting events based on the base rate of occurrences \cite{gervet2020deep}. Therefore, the output values are more reliable when the output calibration is closer to the diagonal line. Overall, the untuned model exhibited poor output calibration; in contrast, output calibration significantly improved as the number of learners in the KTLP dataset used for fine-tuning increased to 16 and 64.

The output of a KT model is the probability of correctly answering an exercise that covers a specific KC. This probability is often regarded as the mastery level of that KC \cite{piech2015deep}. The output of a KT model can also be used for downstream tasks, such as personalized content recommendations \cite{ai2019concept}, making output reliability an essential priority \cite{lee2022contrastive}. The results of this experiment demonstrate enhanced reliability when a model is fine-tuned using KTLP-formatted data.

\subsubsection{4.3 Learning trajectory analysis\\}
To determine whether the method proposed in this study successfully predicts students’ knowledge states, we visually analyzed the students' understanding of each KC in the process of solving exercises. Random selections of students from the science, social studies, and mathematics datasets were made for this purpose. Among several datasets, students for the mathematics subject were selected at random from Assistments09.

Figures 5 to 8 depict the learning trajectories of the students selected for each subject. Heat maps and line graphs were used to visualize how the mastery level for each KC changed as each student progressed through solving problems for each subject.

The x-axis of each heat map represents the interaction (KC id, correctness) at each time step, while the y-axis represents the KC of the questions that the student primarily answered. In other words, the value of each cell indicates the change in the mastery level for the corresponding KC after the interaction occurred. The x-axis of the line plot (lower part of Figures) likewise represents the interaction (KC id, correctness) at each time step, with the y-axis representing mastery level values; thus, each line represents the mastery level of the corresponding KC.


\begin{figure}
\centering
\includegraphics[width=1\textwidth]{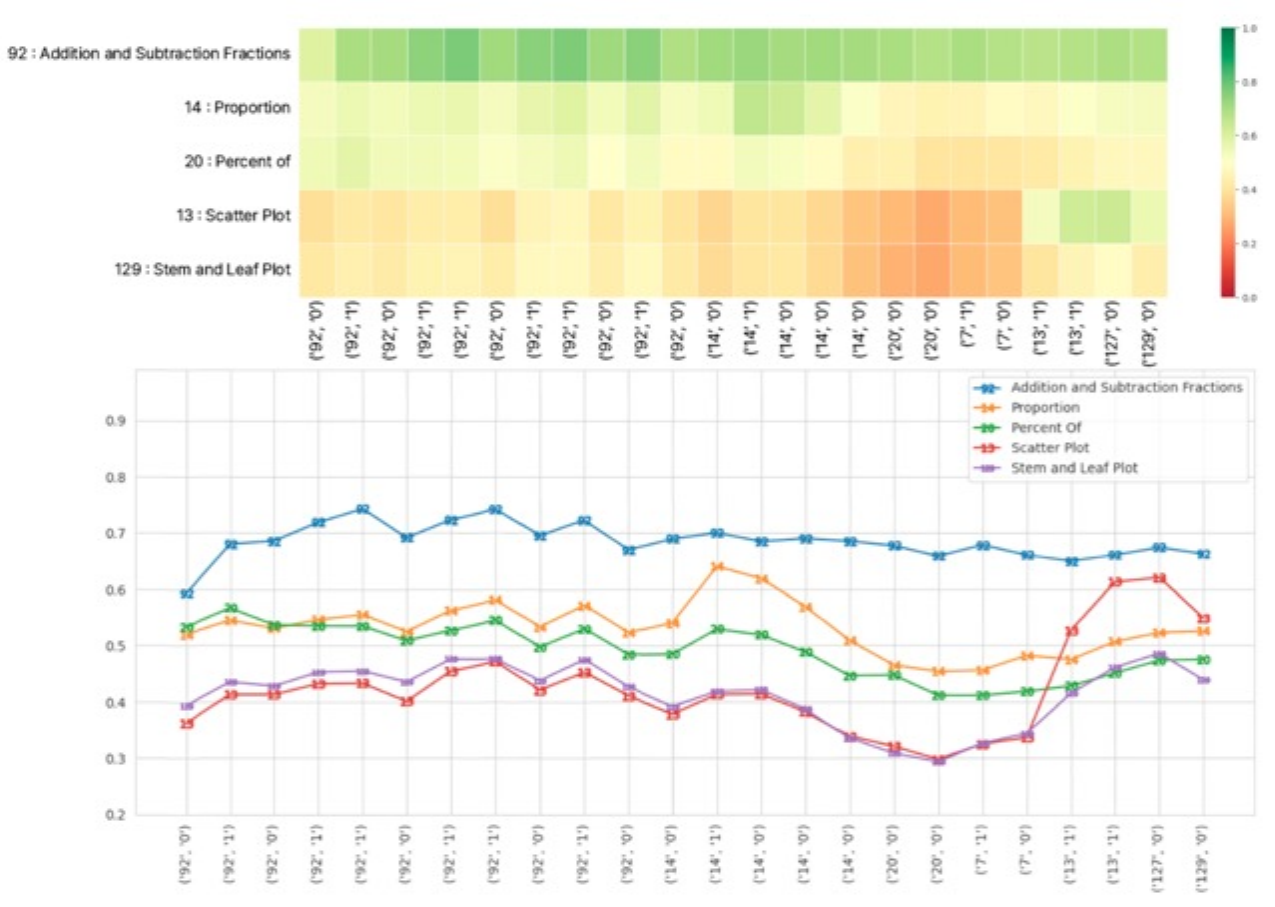}
\caption{Learning trajectories of student 'A' in mathematics} \label{fig5}
\end{figure}

\begin{figure}
\centering
\includegraphics[width=1\textwidth]{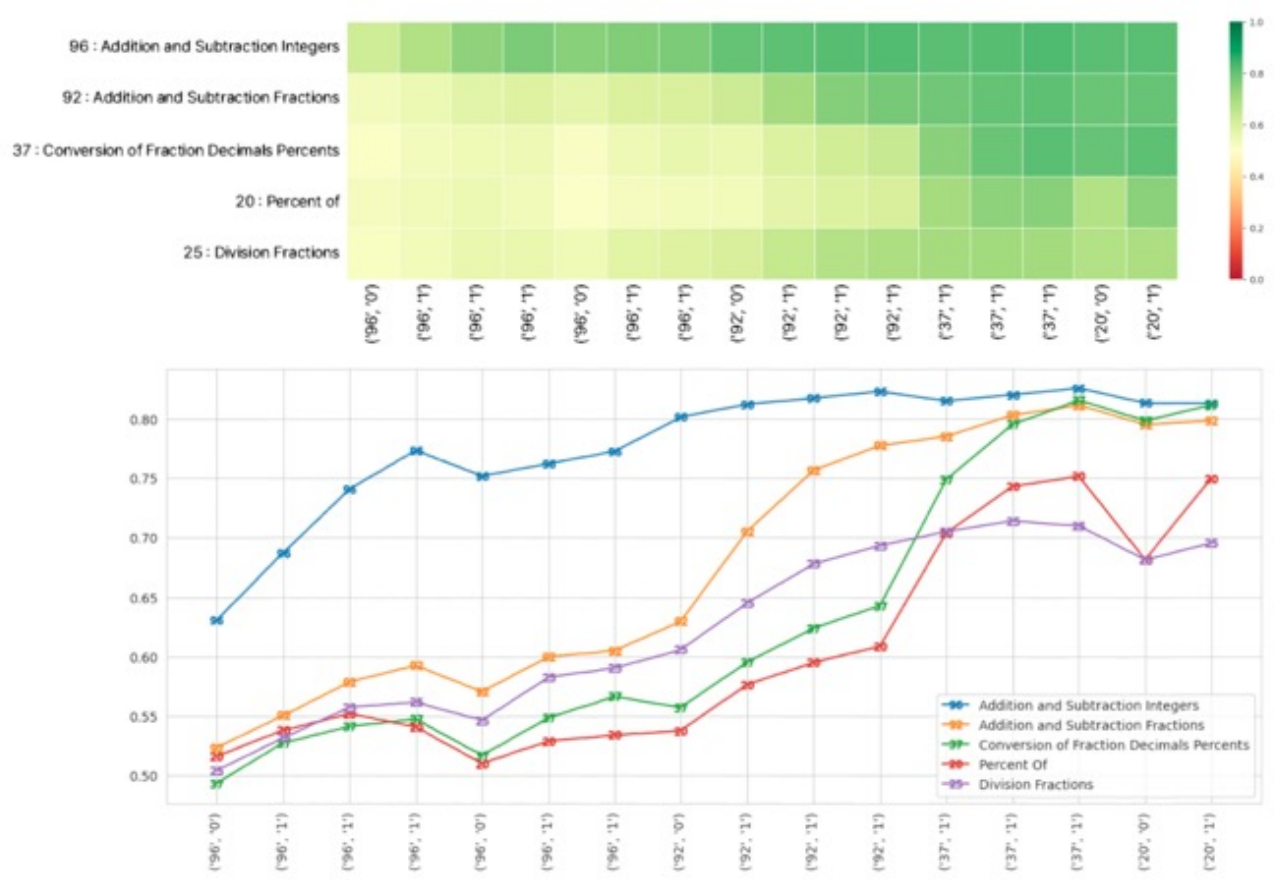}
\caption{Learning trajectories of student 'B' in mathematics} \label{fig6}
\end{figure}

The learning trajectory analysis revealed two observed facts. Initially, each student's knowledge state predicted by CLST was subject to change in accordance with the correctness of the problem. In other words, when a student correctly or incorrectly answers a question about a specific KC, the mastery level of the corresponding KC increases or decreases, respectively. For example, in Figure 5, if the learner correctly answered the question about KC-92, the mastery level for the corresponding KC increased, whereas if the learner answered the question incorrectly, the mastery level decreased. Similarly, when questions about KC-14 were correctly and incorrectly answered, the mastery level of the corresponding KC (orange line in the figure) rose and fell accordingly.

Second, the CLST predicts each learner’s knowledge state by considering relationships between KCs. For example, Figure 5 depicts a math student’s learning trajectory, showing changes in mastery levels for KC-13 (scatter plot), KC-14 (proportion), KC-20 (percent of), KC-129 (stem and leaf plot), and KC-92 (fraction addition and subtraction). Among these, KC-13 and KC-129 are both concepts related to plots, and KC-14 and KC-20 are similarly interconnected. Upon examining changes in the learners' mastery levels, the trends within these pairs of KCs exhibited mutually similar tendencies. In contrast, the mastery level for KC-92 exhibited a different pattern compared to that of the other KCs. 

Furthermore, when a learner correctly responded to an item regarding KC-13, the corresponding mastery level increased along with that of the intuitively related KC-129. Consequently, it is possible to infer that CLST estimates the learner's knowledge state by examining relationships between KCs.

Similarly, in the learning trajectory of another math student shown in Figure 6, the mastery levels of related KCs, namely KC-25 (division of fractions) and KC-92 (addition and subtraction of fractions), as well as KC-20 (percent of) and KC-37 (conversion of fractions, decimals, and percents), displayed similar trends.

In the social studies learning trajectory shown in Figure 7, the mastery level of related concepts KC-2 (national economy) and KC-39 (economic life and choices) changed in a similar manner. For the science learner shown in Figure 8, the mastery levels of related KCs—KC-59 (composition of earth materials) and KC-60 (history of the earth)—changed in similar patterns.

\begin{figure}
\centering
\includegraphics[width=1\textwidth]{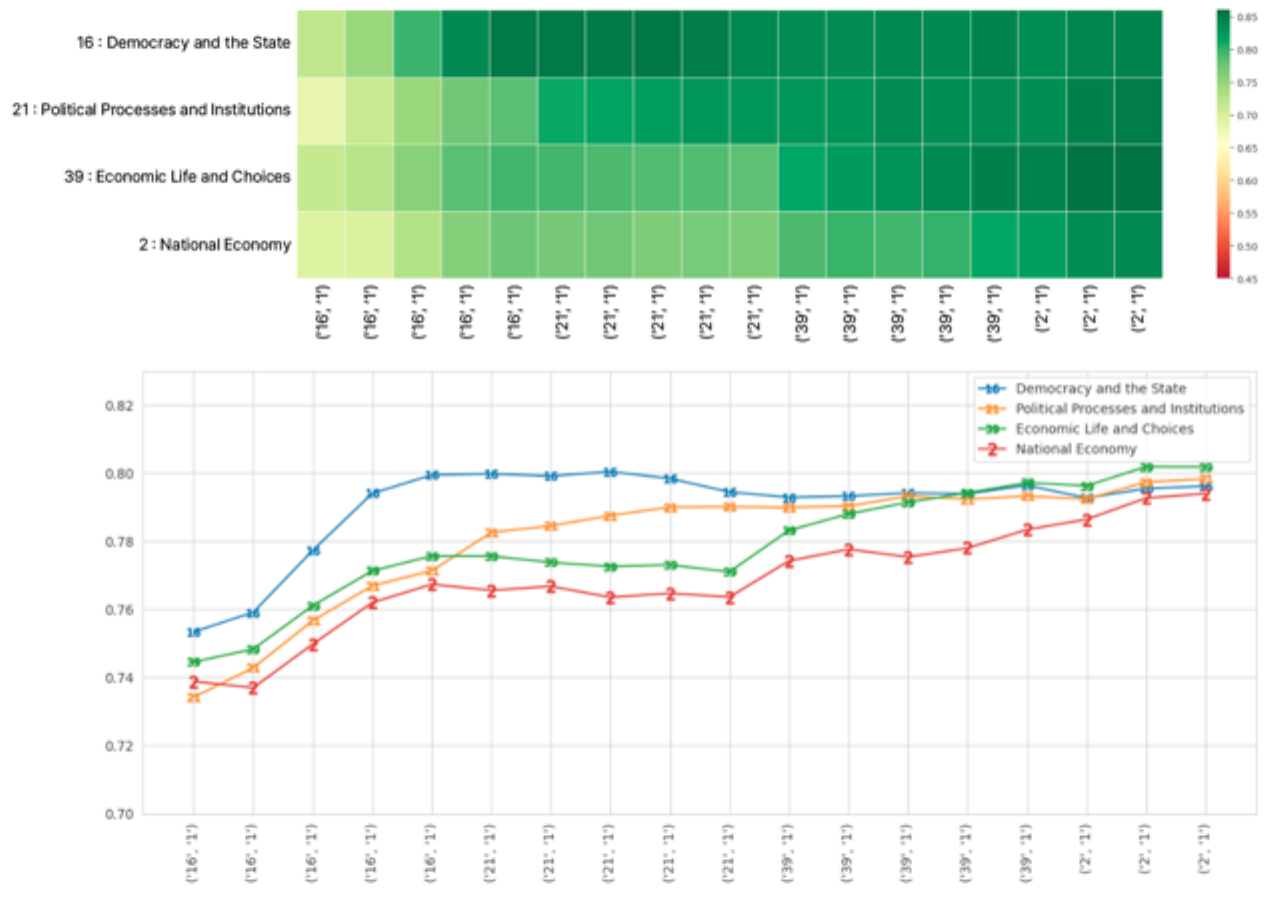}
\caption{Learning trajectories of student 'C' in social studies} \label{fig7}
\end{figure}

\begin{figure}
\centering
\includegraphics[width=1\textwidth]{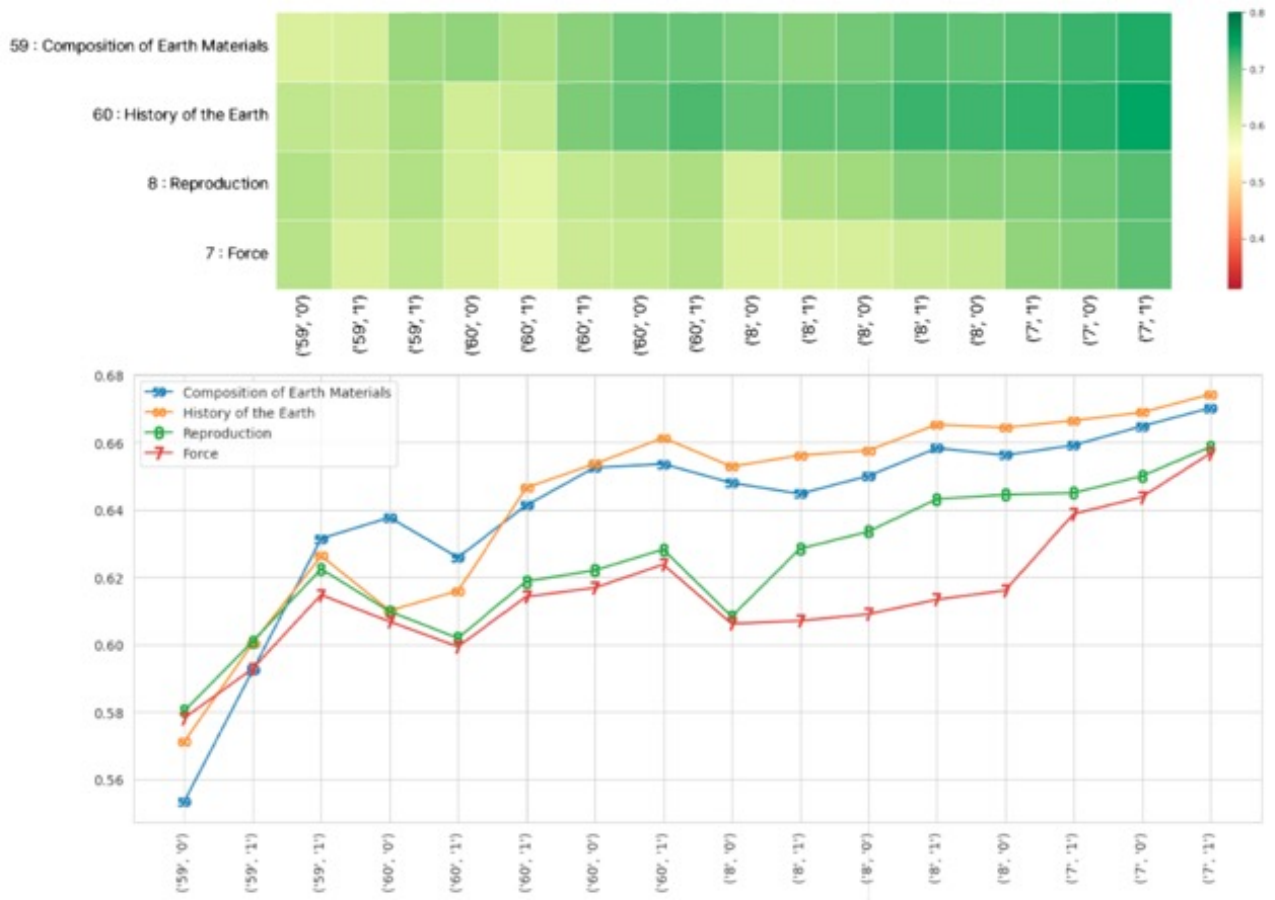}
\caption{Learning trajectories of student 'D' in science} \label{fig8}
\end{figure}

Ultimately, the analysis of learning trajectories yields the following answer to RQ3: The CLST plausibly predicts each student's mastery level and also understands the relationships between KCs.

\subsubsection{4.4 Predictive performance under cross-domain scenarios\\}

Finally, we examined the performance of CLST in a cross-domain scenario.  In this experiment, the terms `domain' and `dataset' are used interchangeably \cite{zhao2019geometry}. To evaluate cross-domain performance, we selected two or more datasets for the same subject. Specifically, we conducted an experiment using the mathematics subject datasets NIPS34, Algebra05, and Assistments09.

We tuned CLST with samples from one source domain and subsequently evaluated its predictive performance on test samples from another target domain. For example, we evaluated the performance of the CLST tuned with samples from the NIPS34 dataset (referred to as CLST(nips) in the figure below) on the Algebra05 dataset. In addition, we evaluated the performance of DL-based KT models and the CLST trained with the target domain samples.

\begin{figure}
  \centering
  \rotatebox{0}{
    \begin{minipage}{1\textwidth}
      \includegraphics[width=1\textwidth]{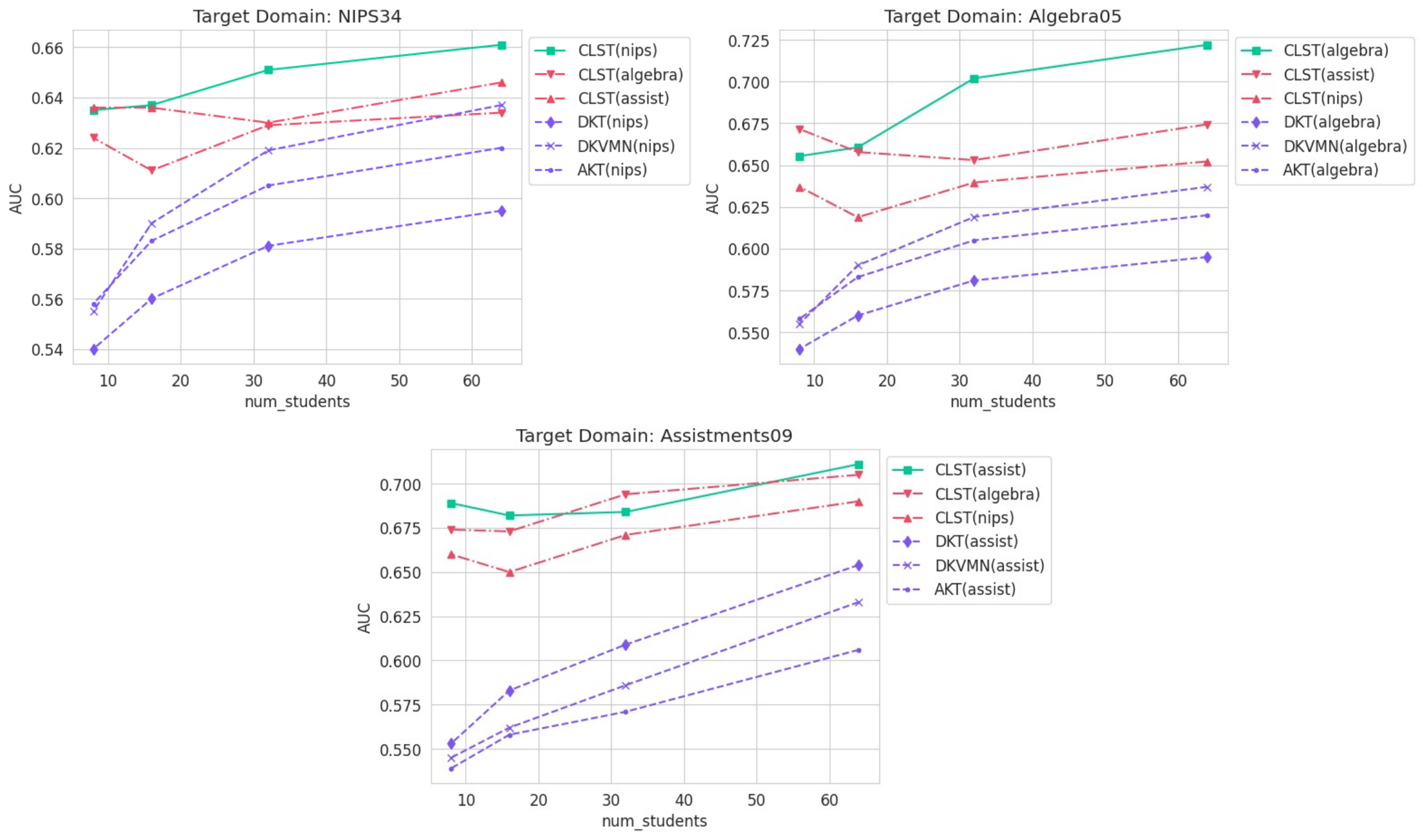}
      \caption{Results of cross-domain experiments.}
    \end{minipage}}
  \label{fig9}
\end{figure}

Figure 9 shows the predictive performance in the cross-domain scenario. According to the overall experimental results, CLST tuned with the target domain data, depicted in green, showed the highest performance. Nonetheless, the CLST trained on a different source domain dataset, denoted by pink dotted lines, outperformed the DL-based KT models trained on the target domain, denoted by purple dotted lines. In particular, in experiments with the Assistments09 dataset, the CLST tuned with different domains (such as algebra and NIPS) demonstrated performance comparable to that tuned with the target domain.

In fact, the majority of current research in the field of KT has primarily concentrated on domain-specific cases. However, real-world scenarios may present obstacles to this approach, such as the inability to obtain data on student–exercise interactions. In such cases, domain-specific models might not function effectively as a result of insufficient data \cite{cheng2022adaptkt}, making the evaluation of cross-domain generalization capacity critical when deploying KT models in practice.

Ultimately, this experiment reveals the following answer to RQ4: Models trained with the CLST are more generalizable across domains; therefore, when building an ITS in situations where data is scarce and sufficient student–exercise interactions cannot be obtained, the CLST can be a promising choice.

\section{Conclusion}
In this study, we developed the CLST, which utilizes a generative LLM as a knowledge tracer. Most existing KT models are designed based on an ID-based approach, which exhibits poor predictive performance in cold-start cases with insufficient data. In contrast, the CLST demonstrates high performance even in cold-start scenarios and exhibits robust cross-domain generalizability. For the experiments, we collected data on the subjects of mathematics, social studies, and science. Subsequently, by expressing the problem-solving data in natural language, we framed the KT task as an NLP task and fine-tuned a generative LLM using the formatted KT dataset.

The experiment results showed that CLST outperformed the baseline models by up to 24.52\% (mathematics), 9.62\% (social studies), and 11.82\% (science) in situations where the number of learners in the training data was insufficient. Furthermore, the ablation study showed that the description-based method applied in CLST, which represents each exercise with its description, achieved better performance than the conventional ID-based method. Additionally, the reliability experiment confirmed that fine-tuning with KTLP-formatted data enhances model reliability. Moreover, the learning trajectory analysis revealed that the CLST plausibly predicts students' mastery levels and understands the relationships between KCs. Finally, the cross-domain KT experiments verified that CLST is generalizable across multiple domains.

The results of this study have practical implications, serving as a guide for educational institutions and EdTech companies interested in fostering personalized learning through the development of ITS. Additionally, it offers theoretical implications by shedding light on strategies for leveraging generative LLMs in the KT domain.

Although the effectiveness of CLST has been demonstrated in numerous experiments, it has the potential to yield even higher predictive performance. For example, the inclusion of additional tasks may increase performance on the target task \cite{wei2021finetuned}. Therefore, it is worthwhile to contemplate future research endeavors to align generative LLMs as superior knowledge tracers by integrating a variety of educational tasks related to KT (e.g., difficulty estimation, KC relation prediction, etc.) into the fine-tuning process.

%
%
%
\bibliographystyle{splncs04}
\bibliography{mybibliography}





\end{document}